\documentclass{article}

\PassOptionsToPackage{numbers, compress}{natbib}

\usepackage[preprint]{neurips_2022}

\usepackage[utf8]{inputenc} 
\usepackage[T1]{fontenc}    
\usepackage{hyperref}       
\usepackage{url}            
\usepackage{booktabs}       
\usepackage{amsfonts}       
\usepackage{nicefrac}       
\usepackage{microtype}      
\usepackage{xcolor}         
\usepackage{graphicx}

\title{Reinforcement Learning with Brain-Inspired Modulation can Improve Adaptation to Environmental Changes}

\author{
 Eric Chalmers\\
Canadian Centre for \\Behavioural Neuroscience\\
  University of Lethbridge\\
  Lethbridge, Alberta, T1K 3M4\\
  \texttt{eric.chalmers@uleth.ca} \\
\And
Artur Luczak \\
Canadian Centre for \\Behavioural Neuroscience\\
  University of Lethbridge\\
  Lethbridge, Alberta, T1K 3M4\\
  \texttt{luczak@uleth.ca} \\  
}

\begin{document}

\maketitle

\begin{abstract}
Developments in reinforcement learning (RL) have allowed algorithms to achieve impressive performance in highly complex, but largely static problems. In contrast, biological learning seems to value efficiency of adaptation to a constantly-changing world. Here we build on a recently-proposed neuronal learning rule that assumes each neuron can optimize its energy balance by predicting its own future activity. That assumption leads to a neuronal learning rule that uses presynaptic input to modulate prediction error. We argue that an analogous RL rule would use action probability to modulate reward prediction error. This modulation makes the agent more sensitive to negative experiences, and more careful in forming preferences. We embed the proposed rule in both tabular and deep-Q-network RL algorithms, and find that it outperforms conventional algorithms in simple, but highly-dynamic tasks. We suggest that the new rule encapsulates a core principle of biological intelligence; an important component for allowing algorithms to adapt to change in a human-like way.
\end{abstract}

\section{Introduction}

\emph{“Most work in biological systems has focused on simple learning problems… where flexibility and ongoing learning are important, similar to real-world learning problems. In contrast, most work in artificial agents has focused on learning a single complex problem in a static environment.”} (Neftci and Averbeck) \citep{neftciReinforcementLearningArtificial2019}

Real-world environments are constantly changing, and the ability to flexibly adapt to these changes is imperative. But current A.I. does not always demonstrate this ability to the same degree as animals. Here, building on a recent model of neuronal learning \citep{luczakNeuronsLearnPredicting2022}, we propose a reinforcement learning rule that demonstrates more realistic flexibility - including both its benefits and its trade-offs. We test the new reinforcement learning rule in multi-armed bandit tasks and a task inspired by the Wisconsin Card Sorting Test - a psychological test used to assess patients’ ability to adapt to changing reward structures. We demonstrate that the new rule improves performance in dynamic decision-making tasks with few to moderate numbers of choices (probably like the routine decision-making faced by animals day-to-day), and that this comes at the expense of performance when selecting between many choices - a paradox-of-choice effect that has been observed in humans. We also discuss some connections between the new rule and several other paradigms from across machine learning.

\section{A new reinforcement learning rule}
\subsection{Basic building blocks of reinforcement learning}

A reinforcement learning agent must be able to estimate the value $V$ of executing action $a$ while in state $s$ - though during the early stages of learning its estimates may not be very good. The agent must learn from each new experience in the environment; improving the efficacy of its value estimates for the future. Suppose at time $t$ the agent is in state $s_t$, executes action $a_t$, and then finds itself in the new state $s_{t+1}$ with reward $r$. The actual, experienced value of this event can be formulated as reward $r_t$ plus the predicted value of being in the new state $s_{t+1}$: 

\begin{equation}
V\left ( s_t, a_t \right )_{actual}=r_t+\gamma V\left ( s_{t+1} \right )
\end{equation}

Here $\gamma$ is a discount factor applied to expected future rewards ($\gamma \epsilon [0, 1]$). The “temporal difference error” $\delta$ expresses the difference between predicted and expected values:

\begin{equation}
\delta_t=V_{actual}-V=r_t+\gamma V \left (s_{t+1} \right ) - V\left ( s_t, a_t \right )
\end{equation}

The temporal difference error is a measure of the agent’s surprise at the recent experience, and is a useful mechanism for learning. In the canonical Q-learning algorithm, for example, the agent maintains a table of value estimates that are updated proportional to $\delta$, and according to a learning rate parameter $\alpha$:

\begin{equation}
V\left ( s_t, a_t \right ) = V\left ( s_t, a_t \right ) + \alpha \delta_t
\end{equation}

The agent selects actions for execution according to a policy $\pi$. For the purpose of this paper, let us assume $\pi$ is a softmax function that calculates the probability of selecting action $a$ out of the set of actions $A$, based on current value estimates, and according to a temperature parameter $\tau$:

\begin{equation}
\pi \left ( s, a \right ) = P\left (a_t=a|s_t=s \right ) = \frac{e^{V\left ( s, a \right ) / \tau}}
{\sum_{b \epsilon A}  e^{V\left ( s, a \right ) / \tau}}
\end{equation}

Thus the learning process consists of iteratively using value estimates to select actions, and using the observed results to improve the value estimates.

\subsection{The new rule}
Building on the Contrastive Hebbian Learning rule \citep{baldiContrastiveLearningNeural1991, almeidaLearningRuleAsynchronous1990, pinedaGeneralizationBackpropagationRecurrent1987}, Scellier and Bengio proposed “Equilibrium Propagation” (EP) as a new, more biologically plausible model for learning in artificial neural networks  \citep{scellierEquilibriumPropagationBridging2017}. EP envisions the network as a dynamical system that learns in two phases. First is the “free phase”, in which an input is applied and the network is allowed to equilibrate. In the second or “weakly clamped” phase, output neurons are soft-clamped or nudged toward a target value. Weights are then updated according to the rule:

\begin{equation}
\Delta W_{ij} \propto \left [ u_i^cu_j^c-u_i^fu_j^f \right ]
\end{equation}

where $i$ and $j$ are the indices of neurons on either side of the weight/synapse (note that EP assumes symmetric connections between neurons), $u^c$ is the neuron’s squashed clamped-phase activation, and $u^f$ is the neuron’s squashed free-phase activation. Luczak, et al. \citep{luczakNeuronsLearnPredicting2022} showed that free phase activity can be well predicted based on past activity, and proposed the following alternative rule:

\begin{equation}
\Delta W_{ij} \propto u_i^c \left ( u_j^c-\tilde{u_j^f} \right )
\end{equation}

Where the tilde indicates the neuron’s prediction of its own free-phase equilibrium given the input. They showed that this rule can explain learning without requiring two distinct phases,  as free phase activity can be predicted in advance. Importantly, the rule arises naturally as a result of a neuron acting to optimize its own energy balance, and hints at an explanation for consciousness \citep{luczakPredictiveNeuronalAdaptation2021}, suggesting that it may encapsulate some principle of general intelligence.  

Examining this new rule, we see the update consists of the prediction error term $\left ( u_j^c-\tilde{u_j^f} \right )$, modulated by the presynaptic activation $u_i^c$. Here we abstract the basic form of this rule to produce a rule applicable to reinforcement learning. The prediction error is easy to place in a reinforcement learning context: it is analogous to the temporal difference error $\delta$. But if we want to formulate a reinforcement learning rule corresponding to the neuronal one, we need a scaling or modulating factor analogous to the presynaptic activation. Since the presynaptic activation is the input to the neuron and the cause of it’s resulting activity, a natural analog could be $\pi \left ( s_t, a_t \right )$; the input to agent’s environment and the cause of the resulting experience. We can then formulate a reinforcement learning rule as a modulation of $\delta$ by $\pi \left ( s_t, a_t \right )$:

\begin{equation}
\Delta V \left ( s_t, a_t \right ) \ \propto \ \pi_t \left ( s_t, a_t \right ) \delta_t \ = \ \pi_t \left ( s_t, a_t \right )\left [ r +\gamma V\left ( s_{t+1} \right ) -V \left ( s_t, a_t \right ) \right ]
\end{equation}

The analogy between the neuronal learning rule and the new RL rule is illustrated in Figure 1. 

\begin{figure}
  \centering
  \includegraphics[scale=0.45]{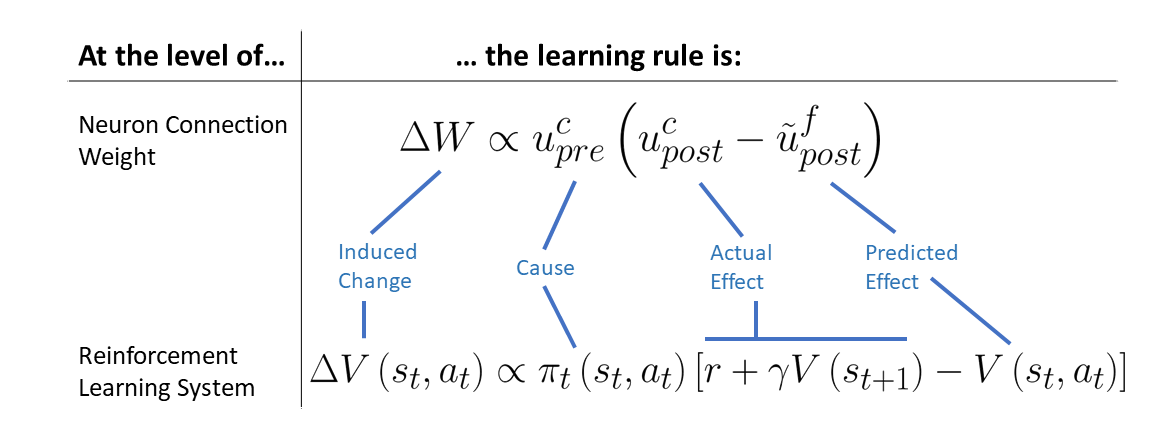}
  \caption{
We abstract the neuronal learning rule of Luczak et al. to create an analogous rule for reinforcement learning. The neuronal rule is based on the assumption that neurons can predict their own future activity. To calculate a weight update, the prediction error is modulated by the presynaptic input that caused the neuron’s activity. In other words, a cause is used to modulate the error of predicting the effect. For an analogous reinforcement learning rule, updates are calculated by using action probability (the cause of the agent’s experience) to modulate reward prediction error (the error in predicting the effect of that action)
}
\end{figure}

\subsection{Effect of the new rule}
Scaling the temporal difference error by $\pi \left ( s_t, a_t \right )$ has two effects:
\begin{enumerate}
\item It magnifies the agent’s reactions to negative experiences. If an action that was thought to be valuable (i.e. $\pi$ is large) brings a negative outcome, the scaled (negative) reward-prediction error will be large. This will depress the perceived value of that action, creating an aversion to it.
\item It slows down the development of an action preference - making the agent somewhat more careful in selecting actions. If the agent is unlikely to take an action, the scaled reward prediction error will be small - even if the experience was rewarding. Thus the agent needs a lot of “convincing” that an unlikely action is actually desirable.
\end{enumerate}

Thus, modulating the temporal difference error by the action probability $\pi \left ( s_t, a_t \right )$ in this way biases the agent somewhat toward negative experiences. We hypothesize this will allow the agent to adapt to environmental changes: when a previously-rewarding action is no longer rewarded, the agent will quickly suppress its perceived value and carefully search for a new preference.

\section{Experiments}
\subsection{Experiment 1 - changing multi-armed bandit}

A changing, n-armed bandit experiment was designed to test the new rule’s ability to adapt to changes. Multi-armed bandits are a simple experiment often used to illustrate learning algorithms’ performance \citep{wangLearningReinforcementLearn2017, duanRLFastReinforcement2016}. The bandit was given one high-reward arm with $p_{reward} = 0.9$ and one no-reward arm with $p_{reward} = 0$. The rest of the arms had random reward probabilities $p_{reward} \sim U(0.25, 0.75)$. The agent receives a +1 reward when the arm it samples is rewarded, and a -1 reward otherwise. The reward probabilities are periodically rotated in such a way that all reward probabilities change, and the arm that was previously high-reward becomes no-reward.

The new rule was implemented in a tabular reinforcement learning agent by modifying the classical Q-learning algorithm to use equation 7 as its update rule. This algorithm maintains a table of the perceived values of each action, and updates the relevant value after each experience. We also implemented a conventional Q-learning algorithm for comparison. These tabular algorithms have no memory and so cannot learn the pattern to the reward probabilities’ rotation. They must respond to each change as if it were a random and complete change to the reward landscape. In our experiments each algorithm was allowed to select the best values for learning rate $\alpha \epsilon [0, 2]$ and softmax temperature $\tau \epsilon {0.5, 1, 2}$. Note that $\alpha$ is conventionally set to be (much) less than 1, but a large value of $\alpha$ can also produce a quick response to environmental changes, so here we allow each algorithm to select $\alpha$ as high as 2. The parameter searches and the experiments themselves were performed on different bandit instances. The cumulative reward for a 7-armed bandit with changes every 100 steps is shown in Figure 2. The conventional Q-learning algorithm outperforms the new rule early on, but the new rule is better able to adapt to the changes, and quickly pulls ahead.

\begin{figure}
  \centering
  \includegraphics[scale=0.75]{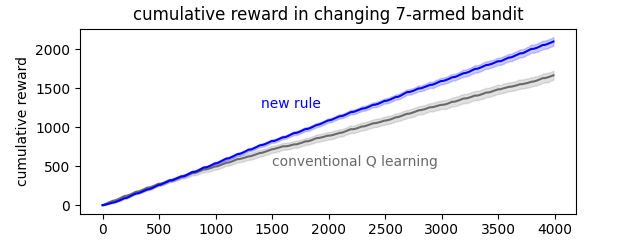}
  \caption{
Cumulative reward obtained on a 7-armed bandit with reward probabilities that change every 100 steps. The shaded area is the 95\% confidence interval of the mean over 20 repetitions. While the conventional Q-learning algorithm is ahead in the early stages of the experiment, the new rule adapts more effectively to the changes over time.
}
\end{figure}

When a change in reward probabilities occurs, the perceived value of the high-reward arm drops sharply for both algorithms. In the new algorithm the probability of selecting the high-reward arm just before the change is very high, and the reward-prediction-error is scaled by this large value. The conventional algorithm achieves a similar sharp drop by self-selecting a large learning rate $\alpha$ (usually somewhere between 0.7 and 1.5). But this large $\alpha$ also causes the conventional algorithm to switch to a new arm very quickly when it finds a chance reward at that arm - sometimes it switches too quickly and selects a sub-optimal arm, and cumulative reward suffers as a result. The new algorithm, on the other hand, scales reward-prediction-error down when the probability of selecting that arm is low, and so spends more time convincing itself that a new arm is desirable. This longer time spent identifying the new high-reward arm yields more reward overall, as shown in Figure 2.

To quantify this extra time taken to develop a new arm preference, we first ran a 10-period moving average on the probabilities of selecting each arm. When the maximum probability of any arm except the previously high-reward arm exceeded 50\%, we considered the agent to have developed a new preference for the corresponding arm. Time-to-preference as the number of bandit arms increases is shown in figure 3.

\begin{figure}
  \centering
  \includegraphics[scale=0.75]{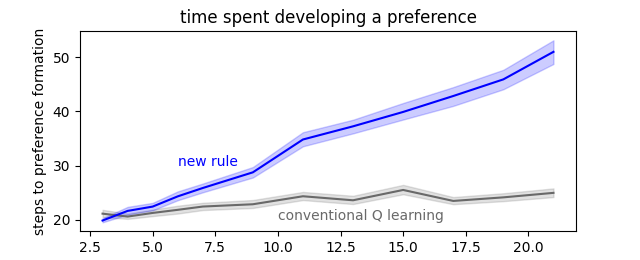}
  \caption{
Time (steps) taken to develop a preference for a new arm after each change, for varying numbers of bandit arms (n). See text for description of how “preferences” were detected. The new rule is more careful in evaluating options, and this helps it to identify the optimal arm when the number of choices is small. The shaded area represents the 95\% confidence interval of the mean over 20 repetitions.
}
\end{figure}

\subsection{Experiment 2 - task inspired by the Wisconsin Card Sorting Test}

The Wisconsin Card Sorting Test is a neuropsychological test used to assess patients’ ability to adapt to a changing set of rules \citep{bergSimpleObjectiveTechnique1948}, and has historically been used to identify brain injury and neurodegenerative disease \citep{milnerEffectsDifferentBrain1963}. The test presents patients with cards that can be matched based on several features, such as color, shape, number, etc. The patients are not told the correct matching criteria, but are rewarded when they make a match correctly. The rewarded matching criteria changes periodically throughout the test: healthy patients can generally adapt quickly when the rule changes, while patients with prefrontal cortex damage cannot.

Here we simulate a similar test using a multiclass classification task. Normally distributed clusters of points are created in n-dimensional space and assigned to each of k classes. The agent is rewarded when it correctly matches a randomly-drawn point to its current class, but the classes are periodically scrambled (such that all the points previously assigned to class “0” now belong to class “2”, for example). 

For this test we use a deep Q network based on the new rule in equation 7. The network is a perceptron with one hidden layer of 20 neurons, and tanh squashing functions. We use separate policy and value networks that synchronize every 5 trials, and a replay buffer of the last 10 trials. The same network is also instantiated with a conventional update rule for comparison. Figure 4 shows the new rule allowing the network to adapt to each change, while the conventional deep Q network adapts less effectively. Here the classification rule is changed every 100 steps.

\begin{figure}
  \centering
  \includegraphics[scale=0.75]{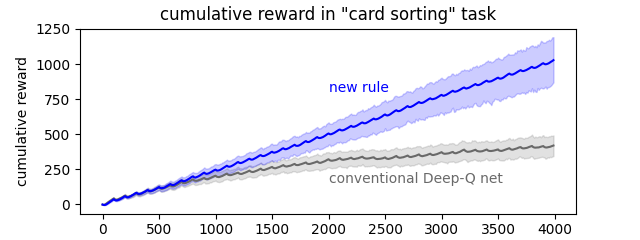}
  \caption{
Cumulative reward obtained in the 4-class version of the task inspired by the Wisconsin Card-Sorting Test, with classes being shuffled every 100 steps. The shaded area is the 95\% confidence interval of the mean over 20 repetitions.
}
\end{figure}

\subsection{The Paradox of Choice}
As humans we often take for granted our ability to change: to update our beliefs in response to new information, or to change a strategy when necessary. But our gift for quick adaptation in everyday situations comes with a trade-off: less-than-optimal performance in situations with many choices. Psychologist Barry Schwartz calls this “the paradox of choice” \citep{schwartzParadoxChoiceWhy2014}: As the number of choices increases, our ability to select a satisfying option decreases and our preferences become weaker \citep{chernevWhenMoreLess2003}.

Our experiments show a similar paradox-of-choice effect, illustrated in Figure 5. The new rule creates a bias toward negative experiences that quickly depresses perceived value of a previously high-reward arm that is now non-rewarding, and also makes the agent more careful in choosing a new preferred arm. This is an advantage when the number of arms is small, but can become a disadvantage if there are many arms. In large-n cases, the agent spends too long evaluating new arms and sometimes fails to select one in time. Thus the same effects of the new rule that allow the agent to effectively navigate environmental changes involving a few choices, hinder it when the number of choices becomes large. This trade-off would likely work heavily in favor of a biological agent, which must continuously adapt to simple though potentially dramatic environmental changes (e.g. the animal’s favorite watering hole now has an alligator in it) in its day-to-day life.

\begin{figure}
  \centering
  \includegraphics[scale=0.75]{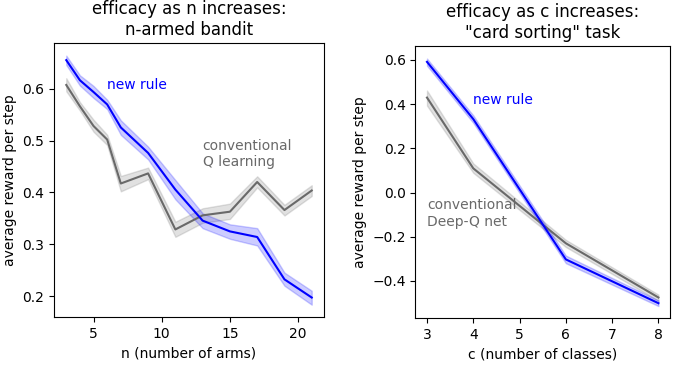}
  \caption{
Average reward-per-step obtained in the n-armed bandit task (left) and "card sorting" task (right). The new rule provides an advantage over conventional learning when the number of choices is small, with the trade-off of a disadvantage when the number of choices is large. This tradeoff is likely favorable for many real-world situations, and similar to the “paradox of choice” effect observed in humans.
}
\end{figure}

\section{Discussion}
The new rule demonstrates some features of human-like learning. Humans are known to increase decision-making time as the number of options increases, in a relationship known as Hick’s law \citep{hickRateGainInformation1952}. Our new rule exhibits similar increasing decision time in Figure 3, while the conventional learning algorithm does not. A paradox-of-choice effect is also observed in Figure 5, where the new rule outperforms conventional learning until the number of choices becomes large. Humans exhibit this trade-off as well, where “selections made from large assortments can lead to weaker preferences” \citep{chernevWhenMoreLess2003} though it should be noted that the relationship between number of choices and the choice-overload effect in humans is complex \citep{chernevChoiceOverloadConceptual2015}. The new rule is derived from a recently proposed neuronal predictive learning rule, and thus may encapsulate some basic principles of learning and intelligence that exist at both the neuronal and system levels. We hope this paper will add to the important conversation around A.I. that can adapt to the constant environmental changes of the real world.

The topics of adaptability and continuous learning represent a growing research field \citep{kwonReinforcementLearningRewardMixing2021, cacciaOnlineFastAdaptation2020, harrisonContinuousMetaLearningTasks2020, liuLearningAdaptEvolving2020}, and paradigms for detecting and responding to environmental change do exist in the machine learning literature. For example, model-based reinforcement-learning approaches maintain an internal model of the world, with which new experiences can be compared to detect environmental changes. Previous work has stored world models and switched between them when recent experiences indicated an environmental change \citep{chalmersContextswitchingAdaptationBraininspired2016a}, adapted time series change-point algorithms to detect environmental changes effectively \citep{padakandlaReinforcementLearningAlgorithm2020}, and used consciousness-inspired approaches to improve the generalization of a model to a new task \citep{zhaoConsciousnessInspiredPlanningAgent2021}. However, this approach requires maintenance of the model, which can be costly. Ultimately the quick, model-free effect of our rule could work well in conjunction with the more complex goal-oriented-planning effect of a model-based approach: the brain employs both model-free and model-based mechanisms \citep{steinkeParallelModelbasedModelfree2020}, and the combination likely holds promise for A.I. as well \citep{asadiatuiStrengthsWeaknessesCombinations2016}.

Another related approach is transfer learning or meta-reinforcement learning, which aims to accelerate learning in new tasks from a previously-experienced family. One meta-RL approach \citep{botvinickReinforcementLearningFast2019} uses a particular recurrent (memory-equipped) network architecture that learns general features of the task family through backpropagation, allowing the recurrent dynamics to quickly tune into details of a new task from the family, in what is thought to be a brain-like mechanism \citep{wangPrefrontalCortexMetareinforcement2018}. Meta-RL is currently an active research field \citep{dorfmanOfflineMetaReinforcement2021, fallahConvergenceTheoryDebiased2021, tangUnifyingGradientEstimators2021}. Unfortunately the network must be informed (reset) each time the task changes, but this general approach could be seen as adaptation through knowledge transfer. Again, the quick memory-free effect provided by our rule could work well in conjunction with such transfer-learning methods, resulting in more human-like learning.

The idea of modulating a prediction error appears elsewhere in machine learning literature, and modulating the error in different ways or by different signals produces different effects. Here we have shown that modulating reward prediction error by action probability creates a human-like adaptation-to-change effect, including improved performance in simple but dynamic tasks, as well as a paradox-of-choice effect. Conversely, the Inverse Propensity Score Estimation (IPSE) approach used in counterfactual learning uses the inverse of the probability as a modulating factor \citep{horvitzGeneralizationSamplingReplacement1952, dudikDoublyRobustPolicy2011}. This can have the effect of de-biasing learning from data collected in a population that differs from a target population. However, during online learning of dynamic tasks it would result in slower adaptation; opposite to our rule. We could also consider REINFORCE-style reinforcement learning algorithms, which modulate a prediction error by a “characteristic eligibility” term that expresses the gradient of the action probability with respect to the parameter being updated \citep{williamsSimpleStatisticalGradientfollowing1992}. This quickly makes rewarding actions more likely - in static environments where the gradient has consistent meaning. Our rule, on the other hand, demonstrates a similar learning effect in dynamic tasks. Making predictions is a central operation of the brain, and it is likely that neural circuits modulate prediction errors in many ways to get the right effect at the right time, creating what we know as human-like learning.

Among the various effects that can be obtained by modulating prediction errors in different ways, we believe the one proposed here deserves special future study for two reasons. First, the ability to cope gracefully in dynamic situations is still relatively understudied (high-profile successes of machine learning are typically in static environments like games). Second, since this new RL rule is derived from a biologically plausible neuronal learning rule, it creates a link between neuron learning and system-level learning which could shed light on universal principles of learning and intelligence.

\section{Code}
Code accompanying this paper can be found at \href{https://github.com/echalmers/modulated_td_error}{https://github.com/echalmers/modulated\_td\_error}.

\bibliographystyle{unsrtnat}

\end{document}